\documentclass{article}
\usepackage{spconf,amsmath,graphicx}
\usepackage{float}
\usepackage{cleveref}

\usepackage{subcaption}
\usepackage{multirow}
\usepackage{url}            
\usepackage{wrapfig}

\title{On using 2D sequence-to-sequence models for speech recognition}
\name{Parnia Bahar$^{1,2}$, Albert Zeyer$^{1,2}$, Ralf Schl{\"u}ter$^{1}$, Hermann Ney$^{1,2}$}
\address{Human Language Technology and Pattern Recognition Group \\
    $^1$Computer Science Department, RWTH Aachen University, D-52056 Aachen, Germany \\
    $^2$AppTek, McLean, USA, \url{http://www.apptek.com/} \\
    { \tt \{bahar, zeyer, schlueter, ney\}@cs.rwth-aachen.de }}

\begin{document}
%
\maketitle
\begin{abstract}

Attention-based sequence-to-sequence models have shown promising results in automatic speech recognition. 
Using these architectures, one-dimensional input and output sequences are related by an attention approach, thereby replacing more explicit alignment processes, like in classical
HMM-based modeling. In contrast, here we apply a novel two-dimensional long short-term memory (2DLSTM) architecture to directly model the input/output relation
between audio/feature vector sequences and word sequences.
The proposed model is an alternative model such that instead of using any type of attention components, we apply a \mbox{2DLSTM} layer to assimilate the context from both input observations and output transcriptions. 
The experimental evaluation on the Switchboard 300h automatic speech recognition task shows word error rates for the 2DLSTM model that are competitive to end-to-end attention-based model.

\end{abstract}
\section{Introduction}
\label{sec:intro}

Conventional automatic speech recognition (ASR) systems using Gaussian mixture model (GMM) and/or hybrid deep neural network (DNN) hidden Markov models (HMM) consist of several components that are trained separately, depend on pretrained alignments and require a complex search  \cite{Hutter_1995_hmm, Robinson_1994_ppe, Bourlard_2012_connectionist, Zeyer_2017_bidi_lstm}. 
Unlike the conventional approaches, attention-based sequence-to-sequence models propose a standalone and single neural network that trains end-to-end, does not need explicit alignments or context-dependent phonetic labels as in HMM and simplify the inference. In these models, an implicit probabilistic notion of alignment is used as part of a neural network. However, it does not work the same way as its analogy of alignment models in the conventional methods.

The widely used attention-based sequence-to-sequence systems are based on an encoder-decoder architecture, where one or more long short-term memory (LSTM) layers read the observation sequence and another LSTM decodes it to a variable length output sequence of characters or words. In such architectures, both input and output sequences are separately handled as a one-dimensional sequence over time.
An attention mechanism is then added into the architecture to combine the encoder and the decoder by allowing the decoder to selectively focus on individual parts of the encoder state sequences \cite{Sutskever_2014_seq2seq, Bahdanau_2015_attention, Chorowski_2015_attention_asr, Bahdanau_2016_asr, Zeyer_2018_attention}.

The LSTM \cite{Hochreiter_1997_lstm} is well suited for sequence modeling, where the sequence is strongly correlated along a one-dimensional time axis. 
Handling dynamic length, encoding positional information, the ability to make use of the previous context and tracking long-term dependencies by the gating strategy are some of the properties which make LSTM appropriate for the sequence to sequence modeling.
Although an LSTM processes essentially one-dimensionally, it can be extended for the processing of multi-dimensional data such as an image or a video \cite{Graves_2008_thesis}. 

In this work, we investigate the use of two-dimensional LSTM (2DLSTM) \cite{Graves_2008_thesis, Graves_2007_mdlstm} in sequence-to-sequence modeling as an alternative model for the attention component. In this architecture, we apply a 2DLSTM on top of a deep bidirectional encoder to relate input and output representations in a 2D space. One dimension of the 2DLSTM processes the input sequence, and another dimension predicts the output (sub)words. In contrast to the attention-based sequence-to-sequence model, where the encoder states are not updated and the model is not able to re-interpret the encoder states while decoding, this model enables the computation of the encoding of the observation sequence as a function of the previously generated transcribed words. Our model is similar to an architecture used in machine translation described in \cite{bahar_2018_2dlstm}. We believe that the 2DLSTM is able to capture necessary monotonic alignments as well as retrieve coverage concepts internally by its cell states. Experimental results on the 300h-Switchboard task show competitive performance compared to an attention-based sequence-to-sequence system.

\begin{figure*}[hbtp]
\centering
\hspace{-0.5cm}
\begin{minipage}{0.45\textwidth}
  \centering
\includegraphics[width=0.95\textwidth]{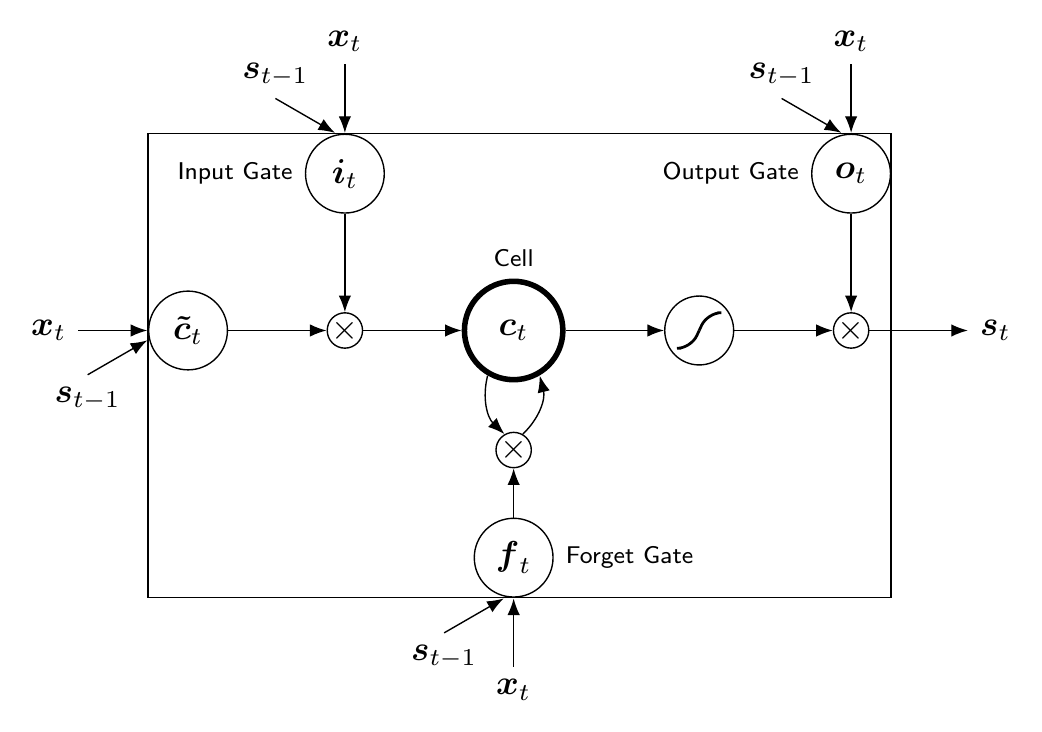}
\subcaption{LSTM}\label{fig:1dlstm}
\end{minipage}%
\begin{minipage}{0.46\textwidth}
  \centering
\includegraphics[width=1.1\textwidth]{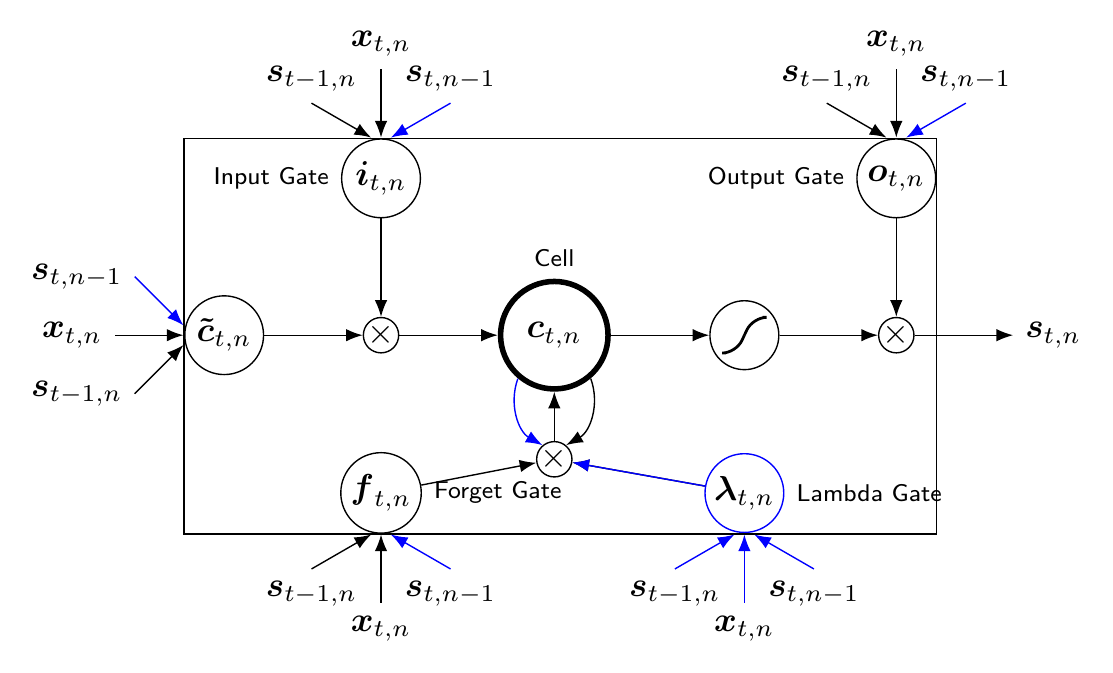}
\subcaption{2DLSTM}\label{fig:2dlstm}
\end{minipage}%
\caption{The internal architecture of the standard and the 2DLSTM. The additional connections are marked in blue \cite{bahar_2018_2dlstm}.} \label{fig:lstm}
\end{figure*}

\section{Related Works}
\label{sec:related_works}

A way of building multidimensional context into recurrent networks is provided by a strategy  that is based on networks with tree-structured update graphs. 
In handwriting recognition (HWR), 2DLSTM has shown successful results in automatic extraction of features from raw 2D-images over convolutional neural networks (CNNs) \cite{Leifert_2016_md_hwr}.
In order to investigate deeper and larger models using 2DLSTM, an algorithm to use the GPU power has been implemented \cite{Voigtlaender_2016_hwr}.

Different neural networks have been proposed in automatic speech recognition (ASR) to model 2D correlations in the input signal. One of them is a 2DLSTM layer which scans the input over both time and frequency jointly for spatio-temporal modeling and aggregates more variations \cite{Li_2016_md_asr}.
Moreover, various architectures to model time-frequency patterns based on deep DNN, CNN, RNN and 2DLSTM layers are compared for large vocabulary ASR \cite{Sainat_2016_md_asr}. 

As an alternative method to the concept of the 2DLSTM, a network of one-dimensional LSTM cells arranged in a multidimensional grid has been introduced \cite{Kalchbrenner_2015_grid_lstm}. In this topology, the LSTM cells communicate not only along time sequence but also between the layers.
The grid LSTM network is also applied for the endpoint detection task in ASR to model both spectral and temporal variations \cite{Li_2017_google_grid_lstm}. 
A 2D attention matrix is also applied in a neural pitch accent recognition model \cite{Bruguier_2018_2d_attention}, in which graphemes are encoded in one dimension and audio frames are encoded in the other.

Recently, the 2DLSTM layer also has been used for sequence-to-sequence modeling in machine translation \cite{bahar_2018_2dlstm} where it implicitly updates the source
representation conditioned on the generated target words. 
In a similar direction, a 2D CNN-based network has been proposed where the positions of the source and the target words define the 2D grid for translation modeling \cite{Elbayad_2018_2d_conv}.

Similar to \cite{bahar_2018_2dlstm}, we apply a 2DLSTM layer to combine the acoustic model (the LSTM encoder) and the language model (the decoder) without any attention components. The 2DLSTM reconciles the context from both the input and the output sequences and re-interprets the encoder states while a new word has been predicted. Compared to \cite{bahar_2018_2dlstm}, our model is much deeper. We use max-pooling to select the most relevant encoder state whereas \cite{bahar_2018_2dlstm} uses the last horizontal state of the 2DLSTM. Furthermore, we utilize the same pretraining scheme
explained in \cite{Zeyer_2018_attention} during training and a faster decoding.

\section{2D Long Short-Term Memory}
\label{sec:background}

The 2DLSTM is characterized as a general form of the standard LSTM \cite{Graves_2008_thesis, Leifert_2016_stable_cells}. It has been proposed to process inherent 2D data of arbitrary lengths, $T$ and $N$. Therefore, it uses both horizontal and vertical recurrences. The building block of both the LSTM and the 2DLSTM are shown in Figure \ref{fig:lstm}. 
At time step $(t,n)$, it gets an input $x_{t,n}$, and its computation relies on both the vertical ${s}_{t, n-1}$ and the horizontal hidden states ${s}_{t-1, n}$.
Besides the input ${i}_{t,n}$, the forget ${f}_{t,n}$ and the output ${o}_{t,n}$ gates that are similar to those in the LSTM, the 2DLSTM employs an additional lambda gate. As written in Equation \ref{mdlstm:lambda}, its activation is computed analogously to the other gates \cite{bahar_2018_2dlstm, Graves_2008_thesis}. 
\begin{align}
{i}_{t,n} &= \sigma \Big( W_{1}{x}_{t,n} + U_{1}{s}_{t-1, n} + V_{1}{s}_{t, n-1}   \Big) \label{mdlstm:1}\\ 
{f}_{t,n} &= \sigma \Big( W_{2}{x}_{t,n} + U_{2}{s}_{t-1, n} + V_{2}{s}_{t, n-1} \Big) \\ 
{o}_{t,n} &= \sigma \Big( W_{3}{x}_{t,n} + U_{3}{s}_{t-1, n} + V_{3}{s}_{t, n-1}  \Big)  \\ 
{\tilde{c}}_{t,n} &= g \Big( W_{4}{x}_{t,n} + U_{4}{s}_{t-1, n} + V_{4}{s}_{t, n-1}  \Big) \\
\label{mdlstm:lambda} 
{\lambda}_{t,n} &= \sigma \Big( W_{5}{x}_{t,n} + U_{5}{s}_{t-1, n} + V_{5}{s}_{t, n-1}  \Big) \\
\label{eq:mdlstm:c}
{c}_{t,n} &=  {f}_{t,n}  \circ \big[ {\lambda}_{t,n}  \circ {c}_{t-1,n} + (1- {\lambda}_{t,n}) \circ {c}_{t,n-1} \big] \nonumber \\
&\phantom{==} +  {\tilde{c}}_{t,n} \circ {i}_{t,n}   \\
{s}_{t,n} &=
g \left( {c}_{t,n} \right) \circ {o}_{t,n}
\end{align}

The internal cell state ${c}_{t,n}$, is computed based on the sum of the two previous
cell's states ${c}_{t-1,n}$ and ${c}_{t,n-1}$, weighted by the lambda gate ${\lambda}_{t,n}$ and its complement (see Equation \ref{eq:mdlstm:c}).
Similar to the LSTM, the internal cell ${c}_{t,n}$ is combined with the output gate to yield the hidden state.
$g$ and $\sigma$ are the hyperbolic tangent and the sigmoid functions. $V_i$, $W_i$ and $U_i$, are the weight matrices. For notational simplicity, we omit the bias vectors.

We process the 2D data in a forward pass from the time step $(1, 1)$ to $(T, N)$ and thus the gradient is passed backwards in an opposite direction from the time step $(T, N)$ to $(1,1)$.
Training a 2DLSTM unit involves back-propagation through two dimensions. For more details, We refer the readers to \cite{Graves_2008_thesis, Leifert_2016_stable_cells}.   

\section{2D Sequence-to-Sequence Model}
\label{sec:2d seq2seq model}

Bayes decision rule requires maximization of the class posterior given an input observation.
In ASR, classes are discrete label sequences of unknown length $N$ (e.g. word, subword, character) sequences, denoted as $w_1^{N}= w_1,\cdots, w_{N}$. 
Given an input observation $x_1^{T}= x_1,\cdots, x_{T}$ of variable length $T$ where usually $T > N$, the posterior probability of a label sequence $w_1^{N}$ is defined as $p(w_1^{N} | x_1^{T})$.
This conditional distribution usually covers the alignment information between the input observation sequence and the output word sequence either implicitly or explicitly.   

In the attention-based sequence-to-sequence approach, the attention weights serve as the implicit probabilistic notion of alignments aligning output labels to encoder states. The freedom of the attention model to focus on the entire input sequence might contradict monotonicity in ASR. In this work, we remove the attention component and intend to investigate whether the 2D sequence-to-sequence modeling is able to properly capture the input-output monotonic relation.

\begin{figure}[htbp]
\includegraphics[width=0.47\textwidth]{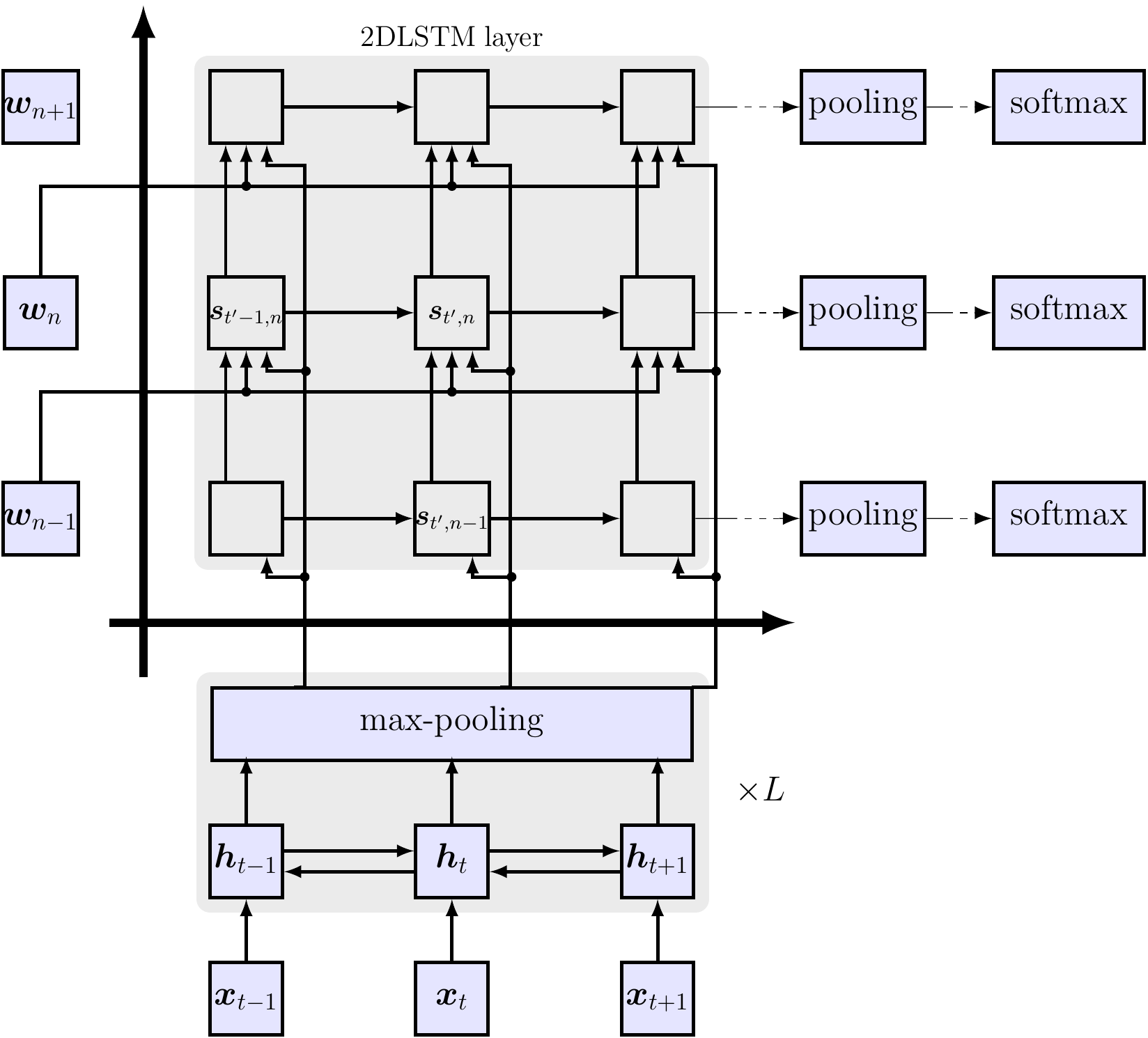}
\caption{The 2D seq2seq architecture using the 2DLSTM layer on top of $L$-layer of encoder. Neither attention components nor explicit LSTM decoders are used. Inspired by \cite{bahar_2018_2dlstm}.} \label{fig:2d_seq2seq}
\end{figure}

As shown in Figure \ref{fig:2d_seq2seq}, we apply a deep bidirectional LSTM encoder ($L=6$) to scan an observation sequence. On top of each bidirectional LSTM layer, we conduct max-pooling over the time dimension to reduce the observation length. Hence, the encoder states are formulated as follows:

\begin{align}
{h}_{1}^{T'} = \text{biLSTM}_{L} \circ \cdots \circ \text{max-pool}_{1} \circ \text{biLSTM}_{1}(x_1^{T})
\end{align}
where $T'$ is the reduced length by a reduction factor. Similar to \cite{bahar_2018_2dlstm}, we then equip the network by a 2DLSTM layer to relate the encoder and the decoder states.
At time step $(t',n)$, the 2DLSTM receives both the encoder state ${h}_{t'}$, and the last target embedding vector ${w}_{n-1}$, as inputs. One dimension of the \mbox{2DLSTM} (horizontal-axis in the figure) sequentially reads the encoder states and another (vertical axis) plays the role of the decoder. Therefore, there is no additional decoder LSTM. Unlike the attention-based sequence-to-sequence model, where the encoder states are obtained once at the beginning, our model repeatedly updates the encoder representations ${h}_{1}^{T'}$, while generating a new output word ${w}_{n}$. 
We note that in this model, we do not use any attention component. 
The state of the 2DLSTM is derived as follows:

\begin{align}
{s}_{t',n} = \text{2DLSTM} \Big(\big[{h}_{t'}; {w}_{n-1}\big], {s}_{t'-1,n}, {s}_{t',n-1}\Big)
\end{align}

It is significant to note that the 2DLSTM state for a label/word step $n$ only have a dependence on the preceding word sequence $w_1^{n-1}$, while it takes into account the whole temporal context of the input observation sequence.

At each decoder step, once the whole input sequence is processed from $1$ to $T'$, we do max-pooling over all horizontal states to obtain the context vector. We have also tried average-pooling or the last horizontal state instead of max-pooling, but none is better in this case.
In order to generate a next output word, ${w}_{n}$, a transformation followed by a softmax operation is applied. Therefore:
\begin{align}
p(w_n| w_1^{n-1}, x_1^{T})  = \text{softmax} \Big( \tanh \big( \text{max-pool} ({s}_{1,n-1}^{T',n-1}) \big) \Big) \big\rvert_{w_n}
\end{align}

\section{Experiments}
\label{sec:experiments}

We have conducted experiments on the Switchboard 300h task. We apply 40-dimensional Gammatone features \cite{Schluter_2007_gammatone} using the RASR feature extractor \cite{Wiesler_2014_rasr}. We use the full Hub5'00 including Switchboard (SWB) and Callhome (CH) as the development set and the Hub5'01 as a test set.
In order to enable an open-vocabulary system, we use byte-pair-encoding (BPE) \cite{Sennrich_2016_bpe} with 1k merge operations.

As our baseline, we utilize the attention-based sequence-to-sequence architecture similar to that described in \cite{Zeyer_2018_attention} with the exact pretraining scheme and the same reduction factor. The baseline model includes a one-layer LSTM decoder with additive attention equipped with fertility feedback. 

The feature vectors are passed into a stack of 6 bidirectional LSTM layers of size 1000 in each direction followed by the max-pooling operation. We downsample the input sequence by factor of 8 in total as described in \cite{Zeyer_2018_attention}. The \mbox{2DLSTM} layer is equipped with 1000 nodes and the output subwords are projected into a 620-dimensional embedding space.
The models are trained end to end using the Adam optimizer \cite{Kingma_2014_adam}, dropout of $30\%$ \cite{Srivastava_2014_dropout}, label smoothing of $0.1$ \cite{Szegedy_2016_label_smoothing} and warmup technique. We reduce the learning rate by a factor of 0.7 following a variant of the Newbob scheme based on the perplexity on the development set for a few checkpoints.

In our training, we use layer-wise pretraining for the encoder, where
we start with two encoder layers and a single max-pool in between with the same multiple-step reduction factor similar to \cite{Zeyer_2018_attention}. 
Decoding is performed using beam search with a beam size of $12$ and the subwords are merged into words. We do not utilize any language model (LM) neither in the baseline system nor in the 2D sequence-to-sequence model.
The model is built using our in-house CUDA implementation of 2DLSTM \cite{Voigtlaender_2016_hwr} 
utilizing optimal speedups in RETURNN \cite{Zeyer_2018_returnn}. The code is open source and the configuration of the setups are available online\footnote{https://github.com/rwth-i6/returnn}. 

Table \ref{tab:ppl_fer} compares the total number of parameters, perplexity and frame error rate (FER) on the development set between our model and the attention baseline. Both models have the same vocabulary size of almost 1K. Our model has 3M more parameters. The perplexity and the FER are comparable. We also compare our model over prior works based on the WER listed in Table \ref{tab:wer}. As a simple significance test, the reported WERs are averaged over 3 runs. Although our 2D sequence-to-sequence model is still behind the hybrid methods, it leads to competitive results over the attention baseline. We observe that our model outperforms the baseline on the Hub5'01 subset by $0.4\%$ absolute. Including a separate LM during the search, we expect to obtain improvements.

\begin{table}[]
\begin{center}
\caption{Total number of parameters, perplexity and FER$^{[\%]}$ on the development set.}
\begin{tabular}{lccc}
\hline
model & \# params & perplexity & FER \\ \hline
baseline \cite{Zeyer_2018_attention} & 157M & 1.56 & 10.9 \\
this work & 160M & 1.53 & 10.6 \\ \hline
\end{tabular}
\label{tab:ppl_fer}
\end{center}
\end{table}

\begin{table}[]
\begin{center}
\caption{WER$^{[\%]}$ on Switchboard 300h. {$^\dag$}average of 3 runs.}
\begin{tabular}{lcccc}
\hline
\multirow{2}{*}{model} & \multirow{2}{*}{LM}  &\multicolumn{2}{c}{\begin{tabular}[c]{@{}c@{}}Hub5'00 \\ (dev)\end{tabular}} & \multicolumn{1}{c}{\begin{tabular}[c]{@{}c@{}}Hub5'01 \\ (test)\end{tabular}} \\
            &           & \multicolumn{1}{c}{SWB}               & \multicolumn{1}{c}{CH}               & \multicolumn{1}{c}{}   \\ \hline
\multicolumn{5}{l}{prior works} \\
\hspace{0.3cm} hybrid  & LSTM & 8.3 & 17.3& 12.9 \\ 
\hspace{0.3cm} CTC \cite{Zweig_2017_advances}  & RNN & 14.0 & 25.3& - \\ 
\hspace{0.3cm} attention \cite{Toshniwal_2017_multitask}  & - & 23.1 & 40.8& - \\ 
\multicolumn{5}{l}{baseline} \\
\hspace{0.3cm} attention \cite{Zeyer_2018_attention}{$^\dag$} & - &13.0 & 26.2& 19.4 \\ \hline
this work{$^\dag$}   &-     &       12.9&        26.4  &     19.0                                                                          \\ \hline
                                                                            
\end{tabular}
\label{tab:wer}

\end{center}
\end{table}

We also compare our model and the attention-based sequence-to-sequence model in terms of decoding speed. Based on the fact that the whole output label sequence is known during the training, the entire 2DLSTM states can be computed once and at each time step, one row of it is taken. This computation cannot be done as a single operation in the search since the output sequence has to be predicted; therefore, during the decoding, we need to compute the states of the 2DLSTM row-wise which slows down the search procedure. This algorithm is faster than \cite{bahar_2018_2dlstm}, where at each output step, they recompute all previous states of 2DLSTM from scratch which are not required. Table \ref{tab:speed} lists the decoding speed of the models to decode the entire development set using a single GPU. In general, the decoding speed of our model is about 6 times slower than that of a standard attention-based model.


\begin{table}[]
\begin{center}
\caption{Decoding speed measured in minutes on the entire development set.}
\begin{tabular}{lc}
\hline
model & decoding speed (mins) \\ \hline
baseline \cite{Zeyer_2018_attention} & 4  \\
this work & 26 \\ \hline
\end{tabular}
\label{tab:speed}
\end{center}
\end{table}

\section{Conclusion}
\label{sec:conclusion}

We have applied a simple 2D sequence-to-sequence model as an alternative to the attention-based model. In our model, a 2DLSTM layer has been utilized to jointly combine the input and the output representations. It processes the observation sequence via the horizontal dimension and generates the output (sub)word sequence through the vertical axis. It does not have any additional LSTM decoder and does not benefit from any attention components. Contrary to the attention-based sequence-to-sequence model, it repeatedly re-encodes the encoder representation when a new output (sub)word is generated. The experimental results are competitive with the baseline on the 300h-Switchboard Hub'00 and show $0.4\%$ improvements on the Hub'01. Our future goal is to develop a bidirectional 2DLSTM to model completely independent of the standard LSTM layers as well as run more experiments on various speech tasks.

\section{Acknowledgements}
\label{sec:acknowledgements}
\begin{wrapfigure}{l}{0.18\textwidth}
\vspace{-4mm}
    \begin{center}
        \includegraphics[width=0.18\textwidth]{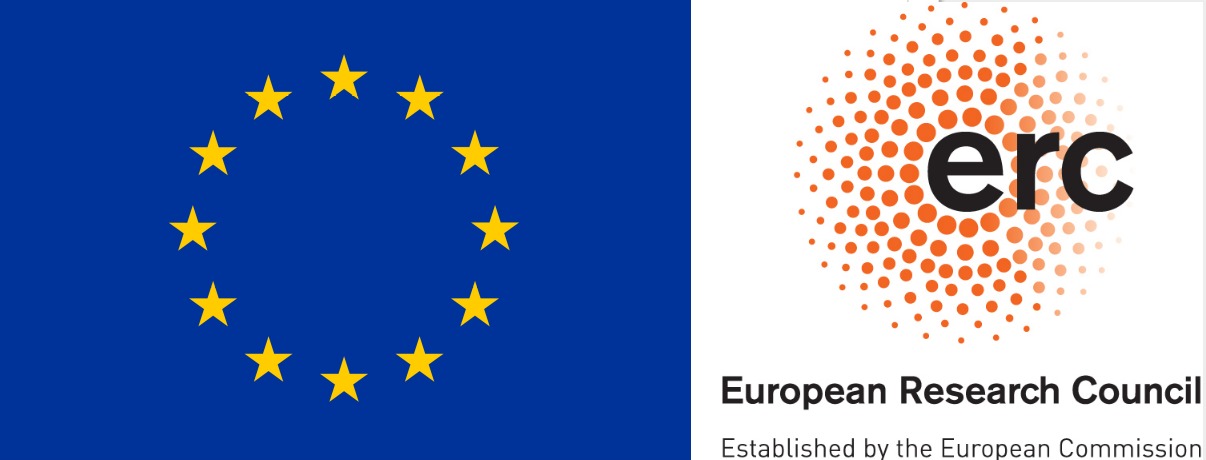} \\
    \end{center}
\vspace{-4mm}
\end{wrapfigure}
This work has received funding from the European Research Council (ERC) under the European Union's Horizon 2020 research and innovation programme (grant agreement No 694537, project "SEQCLAS") and from a Google Focused Award. The work reflects only the authors' views and none of the funding parties is responsible for any use that may be made of the information it contains. 

\bibliographystyle{IEEEbib}


\let\normalsize\small\normalsize

\let\OLDthebibliography\thebibliography
\renewcommand\thebibliography[1]{
  \OLDthebibliography{#1}
  \setlength{\parskip}{0pt}
  \setlength{\itemsep}{0pt plus 0.07ex}
}

\bibliography{main}

\end{document}